%% file: eccv2022submission.tex
\documentclass[runningheads]{llncs}
\usepackage{graphicx}
\usepackage{tikz}
\usepackage{comment}
\usepackage{amsmath,amssymb} % define this before the line numbering.
\usepackage{color}

\usepackage{graphicx}
\usepackage{booktabs}

\usepackage{tabularx, booktabs, ragged2e}
\usepackage{adjustbox}
\usepackage{multirow}
\usepackage{lipsum}  
\usepackage[ruled,vlined]{algorithm2e}

\SetCommentSty{mycommfont}

\def\E{\mathcal{E}}
\def\G{\mathcal{G}}
\def\H{\mathcal{H}}

\def\A{\mathcal{A}}
\def\x{\mathbf{x}}
\def\f{\mathbf{f}}
\def\dsty{\mathbf{d}^{sty}}

\def\dcnt{\mathbf{d}^{cnt}}
\def\w{\mathbf{w}}
\def\m{\mathbf{m}}

\usepackage[pagebackref,breaklinks,colorlinks]{hyperref}
\usepackage[font=small,labelfont=bf]{caption}
\usepackage[caption=false]{subfig}

\usepackage[capitalize]{cleveref}
\crefname{section}{Sec.}{Secs.}
\Crefname{section}{Section}{Sections}
\Crefname{table}{Table}{Tables}
\crefname{table}{Tab.}{Tabs.}

\usepackage[accsupp]{axessibility}  % Improves PDF readability for those with disabilities.

\newcommand{\ie}{\textit{i}.\textit{e}., }
\newcommand{\eg}{\textit{e}.\textit{g}., }
\newcommand{\etal}{\textit{et} \textit{al}.}

\begin{document}
% \renewcommand\thelinenumber{\color[rgb]{0.2,0.5,0.8}\normalfont\sffamily\scriptsize\arabic{linenumber}\color[rgb]{0,0,0}}
% \renewcommand\makeLineNumber {\hss\thelinenumber\ \hspace{6mm} \rlap{\hskip\textwidth\ \hspace{6.5mm}\thelinenumber}}
% \linenumbers
\pagestyle{headings}
\mainmatter
\def\ECCVSubNumber{4718}  % Insert your submission number here

\title{Expanding the Latent Space of StyleGAN for Real Face Editing} % Replace with your title

% INITIAL SUBMISSION 
\begin{comment}
\titlerunning{ECCV-22 submission ID \ECCVSubNumber} 
\authorrunning{ECCV-22 submission ID \ECCVSubNumber} 
\author{Anonymous ECCV submission}
\institute{Paper ID \ECCVSubNumber}
\end{comment}
%******************

% CAMERA READY SUBMISSION
% \begin{comment}
% \titlerunning{Abbreviated paper title}
% If the paper title is too long for the running head, you can set
% an abbreviated paper title here
%
\author{Yu Yin\inst{1} \and
Kamran Ghasedi\inst{2} \and
HsiangTao Wu\inst{2} \and
Jiaolong Yang\inst{2} \and
Xin Tong\inst{2} \and
YUN FU\inst{1}}
\authorrunning{Y. Yin et al.}
% First names are abbreviated in the running head.
% If there are more than two authors, 'et al.' is used.
%
\institute{Northeastern University, USA\\
\email{yin.yu1@northeastern.edu, yunfu@ece.neu.edu} \and
Microsoft \\
\email{kamran.ghasedi@gmail.com, \{musclewu,jiaoyan,xtong\}@microsoft.com}}
% \end{comment}
%******************
\maketitle

%%%%%%%%% ABSTRACT
\begin{abstract}
Recently, a surge of face editing techniques have been proposed to employ the pretrained StyleGAN for semantic manipulation. To successfully edit a real image, one must first convert the input image into StyleGAN’s latent variables. However, it is still challenging to find latent variables, which have the capacity for preserving the appearance of the input subject (\emph{e.g.} identity, lighting, hairstyles) as well as enabling meaningful manipulations. 
In this paper, we present a method to expand the latent space of StyleGAN with additional content features to break down the trade-off between low-distortion and high-editability.
% achieve both identity-preserving and disentangled-attribute editing for real face images.
Specifically, we proposed a two-branch model, where the style branch first tackles the entanglement issue by the sparse manipulation of latent codes, and the content branch then mitigates the distortion issue
by leveraging the content and appearance details from the input image.
% Then the second branch leverages additional texture information from the source image to reduce the editing distortion and generate refined edits with image-specific details well-preserved (\eg, background, appearance, and illumination).
We confirm the effectiveness of our method using extensive qualitative and quantitative experiments on real face editing and reconstruction tasks. 
\end{abstract}

%%%%%%%%% BODY TEXT
\input{sections/introduction}

\input{sections/relatedworks}

\input{sections/method}

\input{sections/experiments}

\vspace{-2mm}
\section{Conclusion}
In this paper, we presented an novel approach to expand the latent space of StyleGAN with additional 2d content features to achieve identity-preserving and attribute-disentangled editing for real face images.
The proposed model includes two different branches to tackle the entanglement and distortion issues in real face editing. The style branch first tackles the entanglement issue by the sparse manipulation of latent codes.
The content branch then queries content and appearance information from the input image to mitigate the distortion issue and refine the edits.
% Qualitative and quantitative evaluation demonstrate the effectiveness of our method.
Qualitative and quantitative evaluation for face reconstruction and editing also demonstrate the effectiveness of our method.

\clearpage
% ---- Bibliography ----
%
% BibTeX users should specify bibliography style 'splncs04'.
% References will then be sorted and formatted in the correct style.
%
\bibliographystyle{splncs04}
\bibliography{egbib}
\end{document}

%% file: sections/introduction.tex
\section{Introduction}
\label{sec:intro}
Generative adversarial networks (GANs)~\cite{goodfellow2020generative} have shown promising performance on image synthesis. In particular, StyleGAN~\cite{karras2019style,karras2020analyzing} is one of the most popular models for face image generative tasks~\cite{jiang2020psgan,men2020controllable,park2020contrastive,chan2021glean}, achieving state-of-the-art results based on visual fidelity and quality. Meanwhile, it has been shown that the latent space of StyleGAN has disentangled properties, providing the opportunity to control different factor of variations in the face image by each of the latent variables. 
%perform semantic manipulation by adjusting the latent codes of the pretrained StyleGAN. 
Several studies~\cite{richardson2021encoding,han2021disentangled,Alaluf2021only} have introduced methods for semantic image editing by manipulating the latent code of StyleGAN, and demonstrated realistic editing of high quality images.
%and powerful synthesizing capabilities of StyleGAN.

%% fig: teaser
\begin{figure}[t]
	\centering   
    \includegraphics[trim=0in 0in 0in 0in,clip,width=\linewidth]{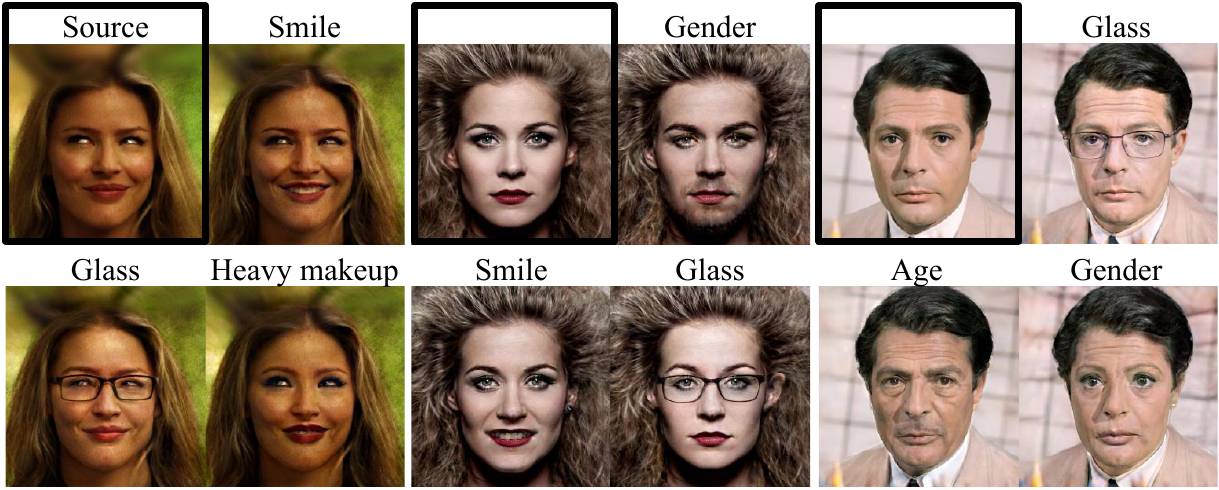}
    \caption{\textbf{Face image editing results of our proposed method.} Our method achieves disentangled semantic manipulation and well-preserved image-specific details  (\eg expression, hairstyles, illumination, and backgrounds). }\label{fig:teaser}
    \vspace{-1mm}
\end{figure}

However, performing meaningful edits on real images is still a challenging problem, since unseen (\emph{i.e.} test) real faces tend to be out of the distributions of StyleGAN latent space in terms of appearance (\eg identities, extreme lighting conditions, and hairstyles). Hence, it is difficult to find a perfect latent representation, which provides the capacity for accurate reconstruction of the input image as well as realistic editing of the image. In fact, some works~\cite{tov2021designing,roich2021pivotal} have shown that StyleGAN-based methods suffer from the trade-off between low-distortion and editability when inverting a real image into the latent space of StyleGAN. It has been reported that editing attributes of a face causes a change in the identity of the subject or some undesired artifacts~\cite{abdal2020image2stylegan++,tov2021designing}.
In addition, performing edits along one attribute may lead to unexpected changes of other attributes, due to the entanglement between different semantics in the StyleGAN latent space.
For example, the "glass" attribute is often entangled with the "age" attribute, which means adding a glass will also change the age of the subject.

In this paper, we present a novel approach to expand the latent space of StyleGAN to achieve identity-preserving and disentangled-attribute editing for real face images. 
In particular, we propose a two-branch method, where the first style branch tackles the entanglement issue by the sparse manipulation of one-dimensional style features, and the second content branch alleviates the distortion issue using two-dimensional content features.
We show that the sparsity constraint on the style features in the first branch is crucial for disentangled and local editing.  
To achieve identity-preserving and low-distortion editing, we leverage additional two-dimensional content features extracted from the input image in the StyleGAN synthesis pipeline. The style and content features are then combined using a feature fusion module in the second branch. 
% Furthermore, an alignment regularization encourages the second branch to refine the edits in some local regions and preserve face-specific details.
Furthermore, an alignment regularization enables the model to identify and maintain non-edited regions to preserve face-specific details.
We show some examples of real face editing by our method in Figure~\ref{fig:teaser}. 
The superiority of the proposed method is demonstrated qualitatively and quantitatively in the reconstruction and semantic editing of real face images. 

The main contributions of this paper can be summarized as:
\begin{itemize}
    \setlength{\itemsep}{0pt}
    \setlength{\parskip}{0pt}
    \setlength{\parsep}{0pt}
    \vspace{-\topsep}
    \item Expanding the latent space of StyleGAN by leveraging two-dimensional content features in the synthesis pipeline to achieve low-distortion editing.
    \item Obtaining disentangled-attribute editing using the sparsity constraint on the style editing directions.
    \item Achieving local region editing using the alignment loss and feature fusion module by steering the effect of style and content features.
    \item Outperforming the state-of-the-art methods on semantic editing and reconstruction of face images using extensive experiments.
    % \item A two-branch method is proposed to tackle the entanglement and distortion of semantic manipulation separately. 
    % \item A regularization term is performed on editing directions to enable disentangled attributes editing.
    % \item We leverage the additional texture information of source images to expand the native latent space of StyleGAN and, hence, achieve realistic edits with image-specific details well-preserved.
    % \item We demonstrate the improved performance of reconstruction and semantic editing on real face images through extensive experiments. 
\end{itemize}

%% file: sections/relatedworks.tex
\section{Related Work}

\subsection{GAN Inversion and Trade-off}
To successfully edit a real image, we need to first invert the image into GAN’s latent space, which is also referred to as \textit{GAN Inversion} in literature.
% Existing inversion approaches can be divided into three categories: (i) optimization-based methods~\cite{abdal2019image2stylegan,karras2020analyzing,creswell2018inverting} directly optimize the latent code from a single image, (ii) encoder-based methods~\cite{alaluf2021restyle,tov2021designing,richardson2021encoding} train a network to estimate the latent code using a large number of images, and (ii) hybrid methods~\cite{guan2020collaborative,zhu2020domain} combine both optimization-based and encoder-based methods to achieve better performance. 
Existing inversion approaches can be divided into three categories: (i) optimization-based methods~\cite{abdal2019image2stylegan,karras2020analyzing,creswell2018inverting}, (ii) encoder-based methods~\cite{alaluf2021restyle,tov2021designing,richardson2021encoding}, and (ii) hybrid methods~\cite{guan2020collaborative,zhu2020domain}. 
In general, encoder-based methods have lower reconstruction accuracy, but are more efficient in editing face images compared to optimization-based methods. In this paper, we mainly focus on encoder-based methods.

Encoder-based inversion method aims to reconstruct the input image with higher accuracy and quality, while ignoring the editability of learned latent codes. Abdal \etal showed that inverting images to the extended space of latent variables $\mathcal{W+}$ achieves better reconstruction compared to the standard $\mathcal{W}$ space of StyleGAN \cite{abdal2020image2stylegan++}. However, Tov \etal~\cite{tov2021designing} pointed out that inverting images to $\mathcal{W+}$ space achieves better reconstruction, but also leads to less editability than $\mathcal{W}$ space. They define this as the trade-off between distortion and editability and design an encoder for better image manipulations. Similarly, the trade-off is also demonstrated in PTI~\cite{roich2021pivotal}, in which an initial inverted latent code is used as a pivot to fine-tune the generator. 
Different from PTI, which needs to be trained and tested on the same set of images, our method can edit unseen images. 
Although PTI has achieved high quality reconstructing and editing results on real images, it has a very high computational cost due to its requirement to fine-tune the generator for each image or a small set of images. Hence, we do not compare it with the proposed method in our experiments.

% Current methods suffer from inversion distortion due to the limitation of information from 1d features...
% We expand the embedding space with 2d features, and make it learn non-attributes editing region. It enables accurate face inversion and disentangled attributes editing

\subsection{Latent Space Manipulation}
Extensive works has explored the latent space of a pretrained GANs for semantic manipulation.
Both supervised~\cite{shen2020interfacegan,abdal2021styleflow,hou2021guidedstyle} and unsupervised~\cite{harkonen2020ganspace,voynov2020unsupervised,shen2021closed} approaches are employed to search for meaningful editing directions.
The supervised approaches (\eg InterfaceGAN~\cite{shen2020interfacegan}, StyleFlow~\cite{abdal2021styleflow}) require supervision using attribute labels, which can be obtained by off-the-shelf attribute classifiers or annotated labels. 
On the other hand, the unsupervised approaches do not find the editing direction based on predefined attributes, but search for meaningful directions according to the distribution of learned latent variables.
%, and then humans visually determine the semantic name after seeing the edited images.
For instance, GANSpace~\cite{harkonen2020ganspace} and SeFa~\cite{shen2021closed} adopt principal components analysis (PCA) and eigenvector decomposition to search for editing directions respectively.

Due to the entanglement among different semantics in the latent space, performing edits along one attribute may lead to unexpected changes of other attributes. To achieve disentangled and local editing, StyleSpace~\cite{wu2021stylespace} proposed to edit in a so-called $\mathcal{S}$ space, changing the effects of a limited number of channels in convolutional features. 
While our method is similar to StyleSpace in the concept of manipulating a sparse set of style features for disentangled attribute editing, our method learns the feature selection process using an end-to-end learning approach, and does not require manual identification of controlling features. Moreover, our method benefits from two-dimensional content features in addition to one-dimensional style features.
% However, in StyleSpace, users must check many different editing directions and manually identify meaningful controls. Inspired by, but different from StyleSpace, we proposed to apply sparse regularization to disentangle directions. 

%% file: sections/method.tex
%% fig: framework
\begin{figure*}[t]
	\centering   
    \includegraphics[trim=0in 0in 0in 0in,clip,width=1\linewidth]{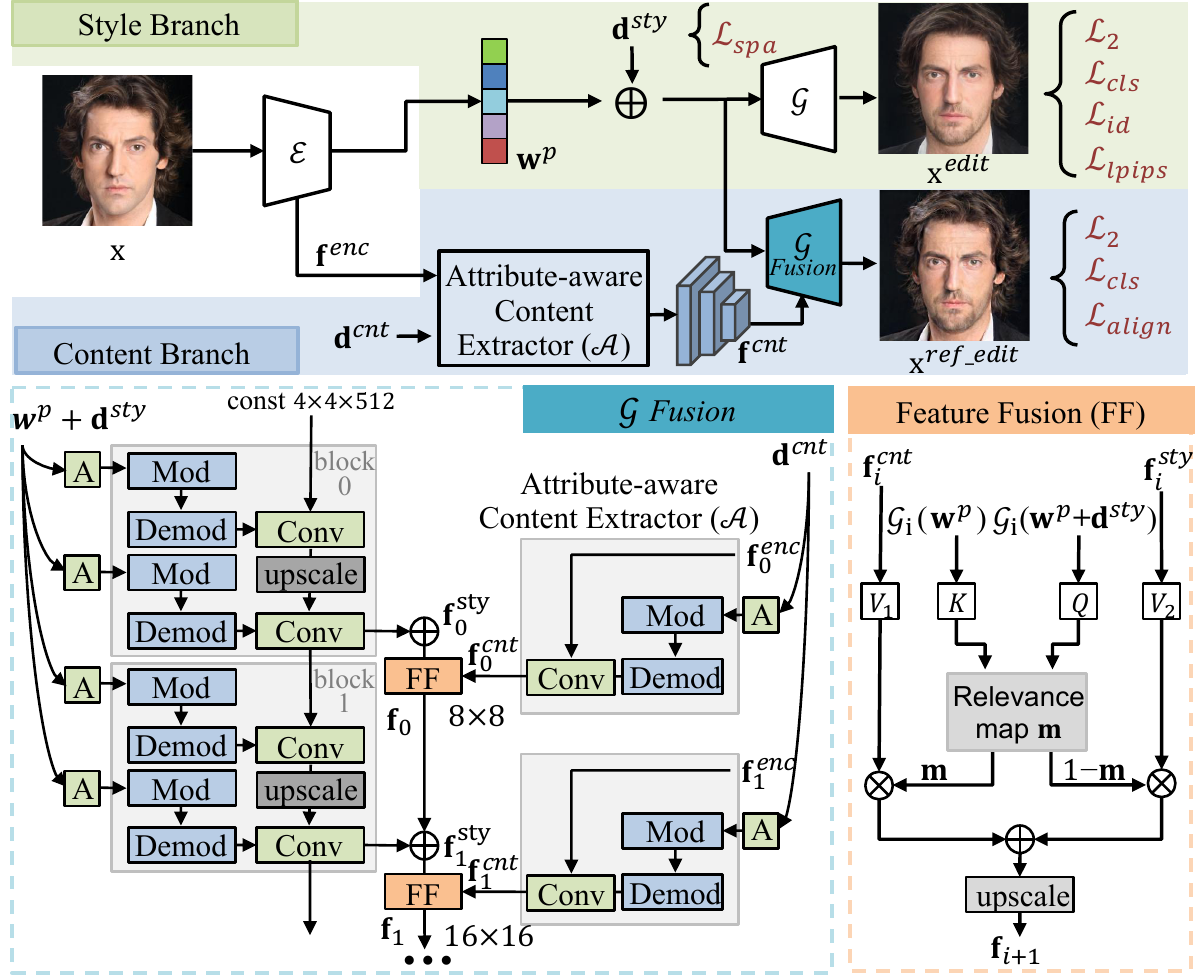}
    \caption{\textbf{The two-branch framework of our proposed method.} 
    The encoder $E$ infers both one-dimensional style features $\textbf{w}^p$ and two-dimensional content features $\textbf{f}^{enc}$ given the input image $\x$. In the style branch, we learn the style editing direction $\textbf{d}^{sty}$ to perform semantic manipulation under the supervision of pretrained attribute classifiers. In the content branch, we learn content direction $\textbf{d}^{cnt}$ to extract attribute-aware content information of the input image and preserve details of the input subject.
    %which will serve as content consultation to refine the edited images. 
    The style and content information are further fused across all layers of $\mathcal{G}$ (bottom left) using an attention-based feature fusion module (bottom right).
    % We further showed the detailed fusion scheme of style and content features (right).
    }
    \label{fig:framework}
    \vspace{-2mm}
\end{figure*}

\section{Methodology}
\label{sec:method}
Our method aims to provide identity-preserving and disentangled-attribute editing for real face images using StyleGAN generator. 
Since the $\mathcal{W+}$ latent space of StyleGAN has the trade-off between the low-distortion and editability, it is very difficult to find latent codes that providing accurate reconstruction as well as retaining powerful editing capabilities. 
To mitigate the distortion-editability trade-off, we propose to expand the latent space of StyleGAN with 2d content features, which provide attribute-aware image-specific details for our editing task.

Given an input face image $\x$ and $K$ target attributes $\mathbf{s} = [s_1, ..., s_K]$, we propose a two-branch framework to synthesize the edited image that have the target attributes while preserving the other appearance details (\ie identity, backgrounds, lighting conditions) of the input images.
In the style branch, we first map the input image into the latent codes of $\mathcal{W+}$ space of StyleGAN using an encoder. Considering a set of learnable directions for the attributes, we then update the latent codes given the editing directions and target attributes. The edited image is generated using the updated latent codes as the input of StyleGAN model.
% synthesis network. 
% A sparsity regularization term is applied to the direction learning to achieve local editing of desired attributes.
In order to improve the fidelity of the edited image, we employ the content branch to enhance the edited image with more appearance details.
%that are not related to edited attributes.
We convey image-specific details through the 2d attribute-aware content features, and combine the style and content features using a feature fusion module.
Figure \ref{fig:framework} shows the general architecture of our proposed method.

\subsection{Sparse Editing of Style Features}
\label{sec:local_editing}
\begin{figure}[t]
	\centering   
    \includegraphics[trim=0in 0in 0in 0in,clip,width=\linewidth]{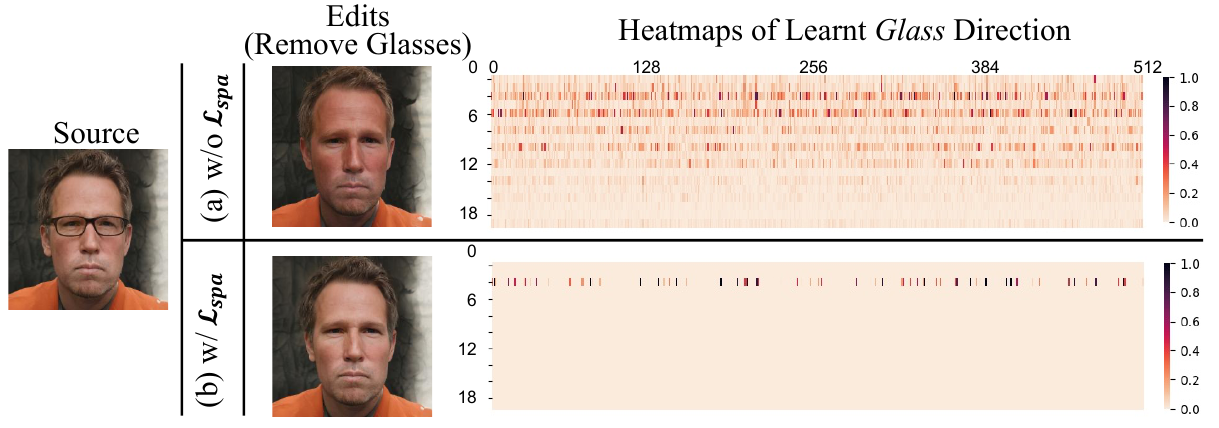}
    \caption{\textbf{Sparsity regularization $\mathbf{\mathcal{L}_{spa}}$ for localized manipulations.} The model is trained with and without $\mathcal{L}_{spa}$, and compared based on the edited image generated by each one. 
    $\mathcal{L}_{spa}$ helps removing unexpected attributes change (\ie the edited image has almost no changes except for the glasses being removed). 
    Furthermore, we show the heatmaps of the normalized \textit{glass} direction with and without $\mathcal{L}_{spa}$. Comparing two heatmaps, it can be seen that the learned direction with $\mathcal{L}_{spa}$ is more sparse, leading to better disentanglement of attributes.
    }\label{fig:glass_dir_heatmap}
\end{figure}

We first convert a real image $\x$ to the latent code $\w^p\in\mathbb{R}^{18\times 512}$ in the latent space of StyleGAN using an encoder $\E$. To perform semantic manipulation on the latent vectors linearly, we learn the style editing directions for each attribute under the supervision of a pretrained attribute classifier.
The editing directions for an image is obtained by a weighted sum of learned directions as follows:
\begin{equation}
\dsty = \sum_{k=1}^K s_k  \dsty_k,
\end{equation}
where $\dsty \in\mathbb{R}^{18\times 512}$ denotes the editing direction for an input image, and ${s_k}$ and $\dsty_k$ show the target variable and the trainable parameters of the $k$-th attribute respectively. 
Then the edit image $\x^{edit}$ can be represented as:
% Then we can edit the input image using the following equation:
\begin{equation}
\x^{edit} = \G(\w^p + \dsty ),
\end{equation}
where $\w^p = \E(\x)$ and $\G$ is the pretrained StyleGAN model.
To encourage disentangled editing, we apply a sparsity regularization on editing directions to reduce the entanglement between different semantics in the latent space. This regularization helps us to perform edits along desired attributes, and avoids unexpected changes corresponding to other attributes.
Moreover, this regularization forces the updated latent variables $\w^p + \dsty$ to be located within the actual distribution of $\mathcal{W+}$ space, leading to generate high-fidelity face images. 
We show the beneficial effect of sparsity regularization in Figure~\ref{fig:glass_dir_heatmap}.
As shown in StyleSpace~\cite{wu2021stylespace}, changing a limited number of channels in the latent space enables disentangled and local editing.
Instead of manually selecting and examining latent variables for each attribute, we found the $\ell_1$-norm loss as a proper candidate for selecting sparse set of latent variables for each attribute, and hence having better control over the visual appearance of local semantic regions. The sparsity regularization can be expressed as: 
\begin{equation}
\mathcal{L}_{spa} =  \sum_k \| \dsty_k \|_1.
\end{equation}

%% fig: glass_dir_heatmap

\subsection{Attribute-aware Content Features}
Due to the inherent distortion of GAN inversion, the edited image $\x^{edit}$ by the style branch has the problem of missing details. To address this issue, we employ an attribute-aware content branch to improve the edited image by preserving high-fidelity details.
The network explicitly consulates the high-frequency texture information from original images as a reference for refining the edited image. 
This sub-network generally consists of two parts, the attribute-aware feature extractor and the feature fusion module.
% the relevance embedding module (RE), the attention-based feature transfer (FT).

% \subsubsection{Attribute-aware Feature Extractor}
\noindent \textbf{Attribute-aware Feature Extractor.}
In order to make the content features aware of editing attributes, we define a new set of editing directions as follows:
\begin{equation}
\dcnt = \sum_{k=1}^K (1 - s_k)  \dcnt_k ,
\end{equation}
where $\dcnt\in\mathbb{R}^{9\times 512}$ is editing direction of content branch, and  $\dcnt_k$ and $s_k$ denote the learnable parameters and target variable for the $k$-th attribute respectively.
The content features $\f^{cnt}$ are extracted using an AdaIN~\cite{karras2020analyzing} based feature extractor as shown in Figure \ref{fig:framework} and the following equation. 
\begin{equation}
\f^{cnt} = \A(\f^{enc}, \dcnt) \,,
\end{equation}
where $\f^{enc}$ denotes the encoder 2d features, and $\A$ represents the feature extractor. 
The details of $\A$ can be found in the Supplementary material. % and ~\cite{karras2020analyzing}.

%% fig: relevanceMap
\begin{figure}[t]
	\centering   
    \includegraphics[trim=0in 0.08in 0in 0in,clip,width=\linewidth]{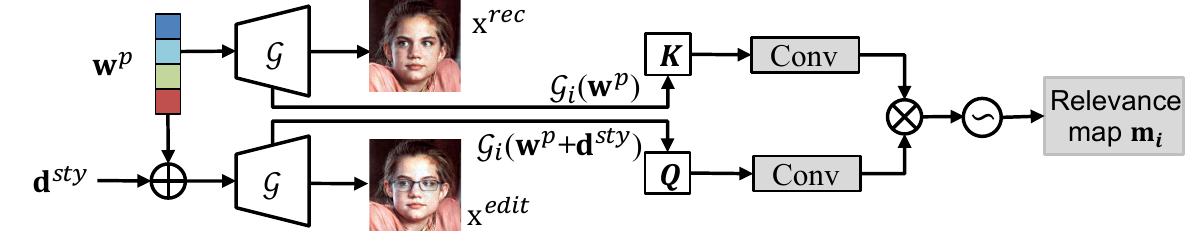}
    \caption{\textbf{Relevance map.} 
    Benefiting from the highly disentangled editing direction $\dsty$, the pixel-wise difference between $\G(\w^p+\dsty)$ and $\G(\w^p)$ should only include local edits. Hence, we compute relevance map by estimating the similarity of the intermediate features $\G_i(\w^p)$ and $\G_i(\w^p+\dsty)$ of the generator $\G$.
    }
    \label{fig:relevanceMap}
\end{figure}

% \subsubsection{Feature Fusion Module}
\noindent\textbf{Feature Fusion Module.}
There are two sets of 2d convolutional features in the  content branch, the style features $\f^{sty}$ obtained from updated latent variables $\w^p + \dsty$ and the content features $\f^{cnt}$ obtained by the attribute-aware feature extractor.  
To fuse the style and content features, we first compute a relevance map $\m_i$ at different layers (\emph{i.e.} resolutions) to find the editing region of interest for each features, and then use an attention module to fuse the features given the relevance map.
%and the other regions that should be intact.
The feature fusion module is shown in Figure \ref{fig:framework}. 

\begin{figure}[t]
	\centering   
    \includegraphics[trim=0in 0.28in 0in 0in,clip,width=0.8\linewidth]{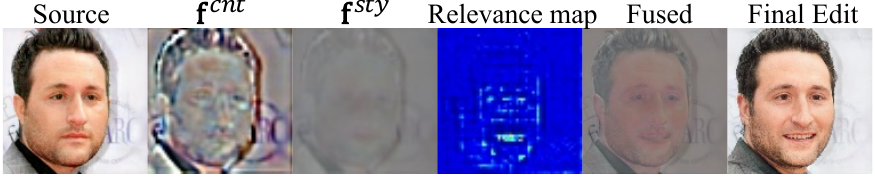}
    \caption{\textbf{Visualization of features at the 5$^{th}$ block of $\G$ \textit{Fusion}.} We show source and edit images in the first and the last column, respectively. In the middle, we show style, content, relevance, and fused features maps in the feature fusion module.}
    \label{fig:mask}
\end{figure}

As shown in Figure \ref{fig:relevanceMap}, relevance map aims to embed the relevance between the style features $\f^{sty}$ and content features $\f^{cnt}$ by estimating the similarity between the reconstructed image $\G(\w^p)$ and the edited image $\G(\w^p+\dsty)$. 
Using the highly disentangled editing direction $\dsty$, the pixel-wise difference between $\G(\w^p+\dsty)$ and $\G(\w^p)$ should only include local edits.
Hence we can infer editing region and refine region using the similarity of the intermediate features $\G_i(\w^p)$ and $\G_i(\w^p+\dsty)$ of the generator $\G$. Then the relevance map $\m_i$ can be computed by:
% Using the similarity of the generated feature $\G_i(\w^p)$ and the edited image $\G_i(\w^p+\dsty)$, we compute the relevance map $\m_i$ using the following equation. 
% Since we learn well disentangled directions in the first branch, the intact region can then be estimated by the similarity between the reconstructed image $G(w_p)$ and the edited image $G(w_p+D^{SE})$ in branch one.
% Then the relevance map $m$ can be computed as:
\begin{eqnarray}
\begin{aligned}
    \m_i 
    % &= sim(K, Q) \\
    &= \H_i \big(  \G_i(\w^p), \G_i(\w^p + \dsty)  \big) \,,
\end{aligned}
\end{eqnarray}
where $\H_i(Q, K) = softmax(Conv(Q)Conv(K)^{\top})$, and $Conv(\cdot)$ is a convolution layer. Intuitively, $\m_i$ can be seen as the region of interest for persevering details of input image (\emph{i.e.} content features), and $(\mathbf{1}-\m_i)$  can be considered as the region for local edits (\emph{i.e.} style features). Therefore, the fused feature in the $(i+1)$-th layer of synthesis network can be expressed as:
\begin{equation}
\f_{i+1} =  (\mathbf{1}-\m_i) \f_i^{sty}  + \m_i \f_i^{cnt} .
\end{equation}
We show an sample visualization of the relevance map and features in Figure~\ref{fig:mask}. 

% \subsubsection{Alignment loss} 
\noindent\textbf{Alignment loss.}
We then propose an alignment regularization that enables the model to identify and maintain non-edited regions, thereby encouraging the model to preserve more face-specific details.
As mentioned in section~\ref{sec:local_editing}, the learned semantic editing direction $\dsty$ is highly disentangled. Hence, the pixel-wise difference between $\G(\w^p+\dsty)$ and $\G(\w^p)$ should only include local edits. Similarly, we aims to refine the edits in the second branch for achieving disentangled editing as well as maintaining high-frequency details of the input image $\x$. In other words, it is desired that the pixel-wise difference between the refined edited image and the input image include only local edits. Thus, we proposed the following alignment loss:
\begin{equation}
\mathcal{L}_{align} = \| \big(G(\w^p+\dsty) - G(\w^p)\big) - \big(\x^{refine} - \x \big) \|_2^2,
\end{equation}
where $\x^{refine}$ is the refined edited image generated by the fusion generator.

% \begin{equation}
% \mathcal{L}_{align} = \sum_{i=0}^{K} ||cos\_sim(d_i, w_{attr}) - s_i)||_2^2 ,
% \end{equation}
% where $K$ is the number of directions.

%% fig: mask

\subsection{Loss function}
The proposed model includes two different branches to tackle the entanglement and distortion of semantic manipulation respectively.
Next we introduce the losses applied on different branches, as well as the overall loss.

% First, we apply a delta-regularization loss to ensure proximity to W when learning the offsets. Second, we use an adversarial loss using our latent discriminator, which encourages each learned style code to lie within the distribution W. 

% \subsubsection{Local editing loss}
\noindent\textbf{Style branch loss. }
The goal of the first branch is to manipulate sparse set of latent codes for disentangled editing, such that editing along an attribute will not cause unexpected changes in the rest of the image.
We employ a pre-trained facial attribute classification model~\cite{han2021IALS} to enable conditional semantic manipulation. Specifically, we estimate attribute labels of the edited image $\x^{edit}$, then calculate the loss with binary cross entropy function. More formally, the attribute classification loss is: % given by
\begin{equation}
\mathcal{L}_{cls} = \frac{1}{K}\sum_{k=1}^K -y_k log (c_k) - (1-y_k)log(1-c_k), 
\end{equation}
where $c_k$ and $y_k$ are the predicted and target labels of the edited image for the $k$-th attribute.

To encourage the disentanglement of semantic edits, we employ a sparsity constrain $\mathcal{L}_{spa}$ introduced in Section \ref{sec:local_editing}. We empirically confirm the effectiveness of this regularization in better disentanglement for local editing. 
Besides, it is also helpful in generating valid faces as it forces the edited latent code to lie within the actual distribution of $\mathcal{W+}$ space.

In addition, we adopt the commonly used $\mathcal{L}_{2}$, $\mathcal{L}_{lpips}$, and $\mathcal{L}_{id}$ losses to enforce pixel-wise, perceptual, and identity similarities between the input and edited images (\emph{i.e.} $\x$ and $\x^{edit}$). Therefore, the loss for the style branch is:
% can be expressed as:
\begin{equation}
    \mathcal{L}_{edit} = \lambda_1\mathcal{L}_{2} + \lambda_2\mathcal{L}_{lpips} + \lambda_3\mathcal{L}_{id} + \lambda_4\mathcal{L}_{cls} + \lambda_5\mathcal{L}_{spa}, 
    % \mathcal{L}_{dist} = \sum_{i} (\mathcal{L}_{2}(x, x^{i}) + \lambda_1\mathcal{L}_{lpips}(x, x^{i}) + \lambda_2\mathcal{L}_{id}(x, x^{i})), 
\end{equation}
where $\lambda_1$, $\lambda_2$, $\lambda_3$, $\lambda_4$, $\lambda_5$ are hyper-parameters of the losses.

% \subsubsection{pixel-level loss}
% L1 loss is used as a pixel-level loss, since it provides better convergence  than L2 in supervised image generation tasks. We adopt pixel-wise L1 loss to measure synthesized faces
% $$L_{1}  = ||x - x_{edit}||_1 $$
% $$L_{lpips}  = p(x) - p(x_{edit}) $$

% \subsubsection{identity-preserving loss}
% Motivated by recent strong self-supervised learning techniques, we generalize this identity loss defined by
% $$L_{id}  = 2 - (f(x) * f(x_{edit})) - (f(x) * f(x_{edit\_refine}))$$
% % where f is a ResNet-50 network and G is the pretrained StyleGAN2 generator. Here, Lsim explicitly encourages the encoder to minimize the cosine similarity between the feature embeddings of the reconstructed image and its source image. Observe that Lsim can be applied in any arbitrary domain due to the general nature of the extracted features. Note that in the facial do- main, we adopt the original identity loss used in pSp, and employ a pretrained ArcFace facial recognition network for extracting the feature embeddings.

% \subsubsection{Editing refinement loss}
\noindent\textbf{Content branch loss. }
The second branch of our model aims to reduce the editing distortion and generates the refined edited image with more appearance details of the input image.

We use the attribute classification loss $\mathcal{L}_{cls}$, the alignment loss $\mathcal{L}_{align}$, and pixel-wise loss $\mathcal{L}_{2}$ to train the second branch. These losses assist us to have less distortion and local editing in the refined edited image (\emph{i.e.} $\x^{refine}$). 
% $\mathcal{L}_{2}$ forces the model to rely more on semantic latent code,  forces the model to rely more on 2d content features.
% $\mathcal{L}_{2}$ and $\mathcal{L}_{cls}$ compete with each other to the texture consultation direction $\dcnt$.
Note that sparsity regularization is employed only on the style editing direction $\dsty$, but not the content direction $\dcnt$.
\begin{equation}
    \mathcal{L}_{refine} = \lambda_6\mathcal{L}_{2} +  \lambda_7\mathcal{L}_{cls} +
    \lambda_8\mathcal{L}_{align},
\end{equation}
where $\lambda_6$, $\lambda_7$, $\lambda_8$ are hyper-parameters that control the trade-off between the loss functions. 

% \subsubsection{Overall loss}
\noindent\textbf{Overall loss. }
The following shows the overall objective function of our model:
\begin{equation}
    \mathcal{L} = \mathcal{L}_{edit} + \mathcal{L}_{refine}.
\end{equation}
% where $\lambda_1$, $\lambda_2$, $\lambda_3$, $\lambda_4$, and $\lambda_5$ are hyper-parameters that control the trade-off of the loss terms. 
% A total variation regularization $\mathcal{L}_{tv}$ \cite{johnson2016perceptual} is also included to remove unfavorable artifacts in synthesized frontal faces $I^{SF}$.

% \subsubsection{Cycle loss}
% $$
% L_{cyc} = ||w - E(G(w))||_2^2,
% $$
% where $w = [w_{attr}; w_{id}]$.

% \subsubsection{Swap loss}
% $$
% L_{swap} = ||w_{attr} - E_{attr}(G(w_{attr},  w^\prime_{id}))||_2^2 + ||w_{id} - E_{id}(G(w^\prime_{attr},  w_{id}))||_2^2,
% $$

%% file: sections/experiments.tex
\section{Experiments}\label{sec:exp}

In this section, we comprehensively evaluate our method by comparing it with the state-of-the-art inversion methods in terms of real face editing and reconstruction. 
Both qualitative and quantitative results are reported.
We then conduct ablation studies as a deep-dive revealing the contributions of the components introduced in this work, as well as the limitations of the work.

\subsection{Datasets and Implementation Details}
% \subsection{Implementation Details}
\noindent\textbf{Datasets. }
FFHQ~\cite{karras2019style} and CelebA-HQ~\cite{karras2017progressive} are two high-quality face datasets with a wide range of attribute variance (\eg gender, age, smiling, glass).
In our experiments, we use the FFHQ dataset~\cite{karras2019style} as training set, and evaluate on the test set of CelebA-HQ~\cite{karras2017progressive}. 

\noindent\textbf{Implementation Details. }
We use a pretrained StyleGAN2 model~\cite{karras2020analyzing} as the generator, and used pSp~\cite{richardson2021encoding} or e4e~\cite{tov2021designing} model as the encoder in our experiments. 
We employ facial attribute classifiers to provides the supervision for conditional semantic manipulation.
The classifiers are ResNet-18~\cite{he2016deep} pretrained on the CelebA~\cite{liu2018large} dataset.
We also employ a pretrained face recognition model (\ie ArcFace~\cite{deng2019arcface}) to enhance the identity preserving ability of the model.
% To enhance the identity preserving ability of the model, we employ a pretrained face recognition model (\ie ArcFace~\cite{deng2019arcface}) to extract meaningful feature representations and minimize the identity similarity between source and edited images.
% The output image with the resolution of 1024$\times$1024 are resized to 256$\times$256 before being fed into the loss functions.
% For training, we used an ADAM optimizer with a learning rate of $10^{-4}$. The model is trained for 80K iterations with batch of 16. 
Hyper-parameters were set as follows: $\lambda_{1}=0.01$, $\lambda_{2}=0.8$, $\lambda_3=1$, $\lambda_4=10$, $\lambda_5=1$, $\lambda_6=0.01$, $\lambda_7=10$, $\lambda_8=1$.
More implementation details can be found in the Supplementary Material.

%% fig: SOTA_comparison
\begin{figure*}[t]
	\centering   
    \includegraphics[trim=0.2in 0in 0in 0in,clip,width=\linewidth]{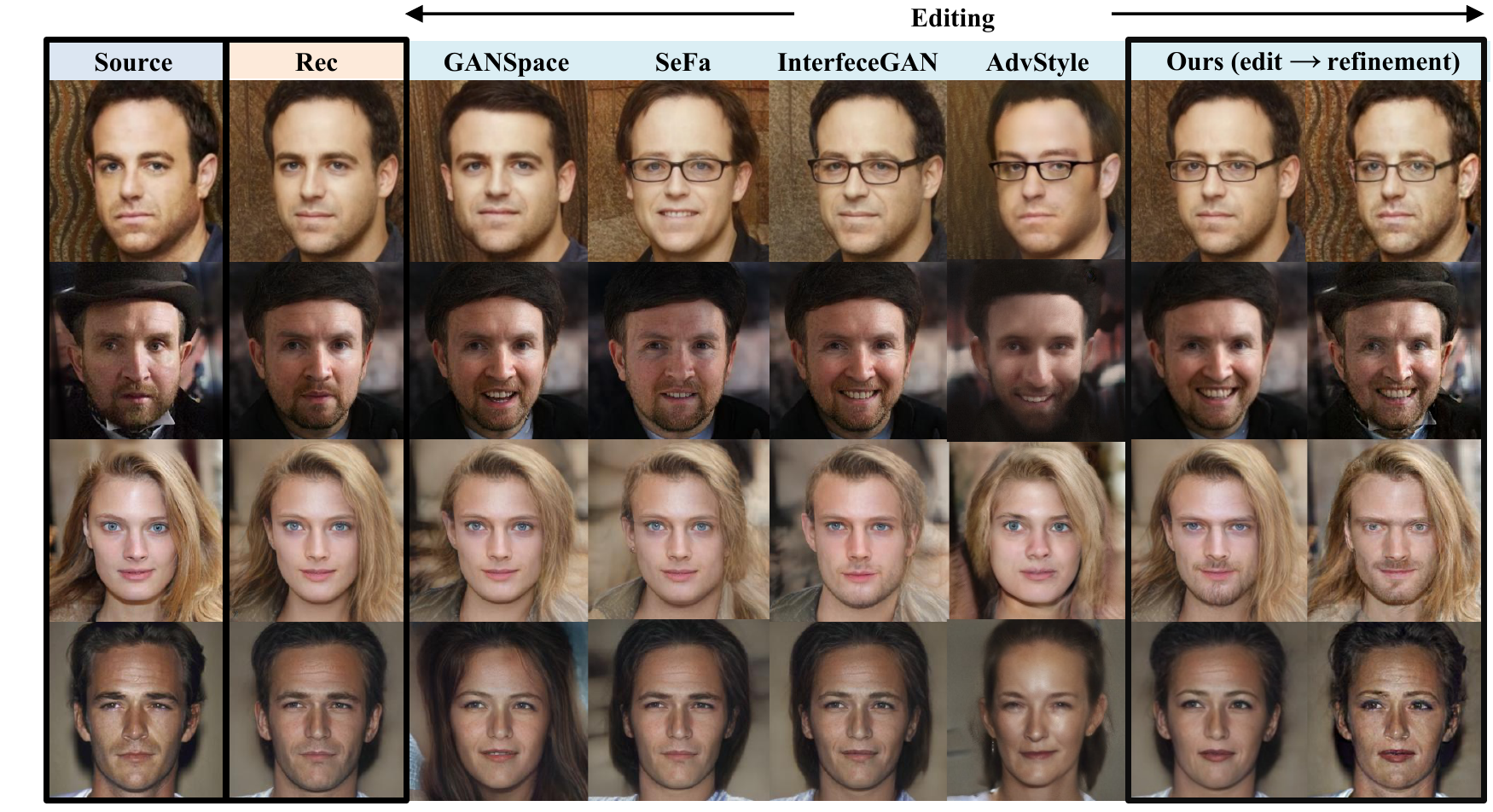}
    \caption{\textbf{Editing comparison of images from the CelebA-HQ dataset.} We demonstrate glass, smile and gender (the last two rows) edits. To edit a real face image, we first invert the image into $\mathcal{W+}$ space using encoders e4e~\cite{tov2021designing} (the first two rows) and pSp~\cite{richardson2021encoding} (the last two rows), then employ the supervised (\ie InterfaceGAN~\cite{shen2020interfacegan}, AdvStyle~\cite{yang2021discovering}) and unsupervised editing methods (\ie GANSpace\cite{harkonen2020ganspace}, and SeFa~\cite{shen2021closed}) on the estimated latent codes from both encoders. 
    % As can be seen, faces reconstructed by e4e has larger distortion, but better editability. Compared to e4e, pSp has better reconstruction (\ie accurate hairline and facial expressions), but worse editability.
    Comparing to different editing methods applied on either e4e or pSp latent codes, our method achieves significantly better attribute editing with image-specific details well-preserved (\eg, back-ground, appearance and illumination). Besides, we also show that our method encourages local and disentangled manipulation of desired attributes, while other methods may include some unexpected changes of other semantics.}\label{fig:SOTA_comparison}
    \vspace{-2mm}
\end{figure*}

%% fig: showcase
\begin{figure*}[t]
	\centering   
    \includegraphics[trim=0in 0in 0in 0in,clip,width=\linewidth]{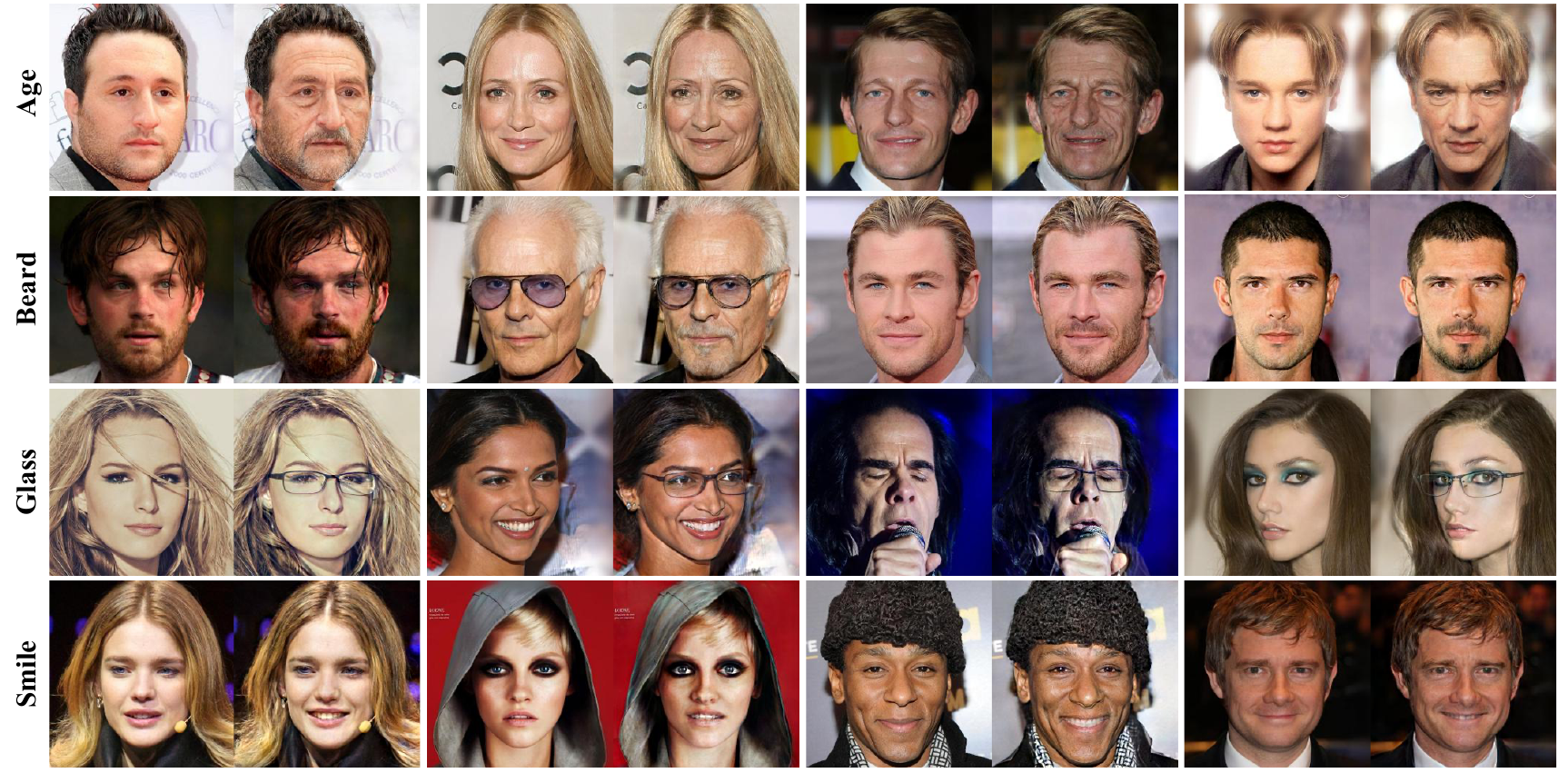}
    \caption{\textbf{Sample results of manipulating real faces along various attributes.} We demonstrate the following edits (top row to bottom): age, beard, glass, smile. The edit of aging mainly adds realistic wrinkles and whitens the beard on faces. As shown in the second row, the type of beard added is adaptive to the source image. Glass editing can be performed on challenging cases (\eg partial occlusion and large poses). Smile is well disentangled with other attributes.
    % Our method is able to preserve all other details (\eg accessories, hats, and hairstyles) and only manipulate the smile on the face.
    }\label{fig:showcase}
    % \vspace{-4mm}
\end{figure*}

\subsection{Semantic Face Editing}

In this section, we demonstrate the superior performance achieved by the proposed method on semantic face editing.
Specifically, we estimate latent codes using two different encoders (\ie e4e~\cite{tov2021designing} and pSp~\cite{richardson2021encoding}), then perform semantic manipulation using various editing methods. Editing results are evaluated qualitatively and quantitatively.

% \subsubsection{Qualitative evaluation}
\noindent\textbf{Qualitative evaluation.}
We show comparison with different editing methods including InterfaceGAN~\cite{shen2020interfacegan}, AdvStyle~\cite{yang2021discovering}, GANSpace\cite{harkonen2020ganspace}, and SeFa~\cite{shen2021closed}.
Note that unsupervised methods (\eg GANSpace, SeFa) might fail to edit unnatural attributes (\ie glass) due to the lack of supervision from attribute labels. In contrast, supervised methods (\eg InterfaceGAN, AdvStyle, and ours) generally work well on those attributes.

As shown in the first two rows of Figure~\ref{fig:SOTA_comparison}, we invert images into $\mathcal{W+}$ space using the encoder e4e~\cite{tov2021designing}, then compare the quality of edited faces generated using different latent-based editing methods.
We observe large distortions in the reconstructed image, such as minor identity changes, inaccurate hairstyles, changed facial components (\eg shape and color of lips). However, all editing approaches based on the e4e encoder can achieve meaningful and visually pleasing semantic manipulations.
Comparing to these methods, our method not only enables disentangled style editing, but also refines the editing results with more content and texture details of the source image.

The last two rows of Figure~\ref{fig:SOTA_comparison} shows the editing comparisons based on the latent code estimated from the pSp~\cite{richardson2021encoding} encoder.
Though pSp can provide better reconstruction (\eg accurate hairline and facial expressions), it suffers from poor editability.
Comparing to different editing approaches applied on the pSp latent codes, our method shows better capability for disentangled manipulation along a single attribute. Moreover, it can be seen that our method is the only one that can preserve a large amount of details in the refined results.

Furthermore, we show in Figure~\ref{fig:showcase} that our approach achieves impressive editing results along different attributes. More results on various attributes can be found in the Supplementary Material.
% This proves that current inverting-editing methods suffer from the distortion-editability tradeoff.
% Note that pSp~\cite{richardson2021encoding} has better reconstruction, but worse editability, while e4e~\cite{tov2021designing} sacrifices reconstruction quality for better editability.

%% fig: reconstruction
\begin{figure}[t]
	\centering   
    \includegraphics[trim=0in 0in 0in 0in,clip,width=0.85\linewidth]{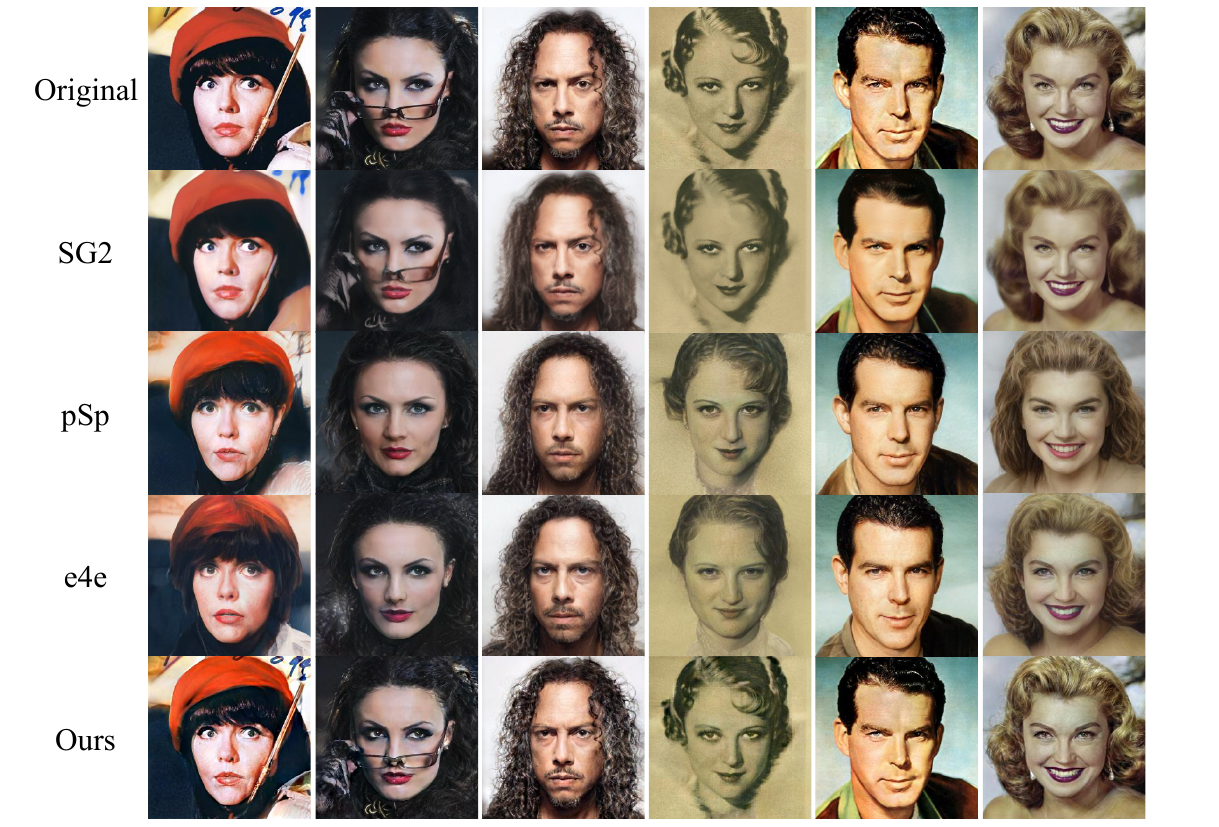}
    \caption{\textbf{Reconstruction quality comparison. } Our method can reconstruct out-of-domain samples very well (\ie the first two rows), while other methods fail to reconstruct due to blurry, wrong or completely face accessories (\ie hat, glasses). 
    For less challenging inputs (\ie the last two rows), we show that the proposed method achieves better reconstruction quality with more accurate details (\ie hair style, cloth, eyebrows) than the other methods. 
    Zoom-in recommended.
    }\label{fig:reconstruction}
\end{figure}

%% fig: reconstruction
%% tab: editing
%% tab: reconstruction
\begin{figure}[t]
\begin{minipage}{\linewidth}
%   \begin{minipage}[c]{0.48\textwidth}
%     \centering
%     \includegraphics[trim=0in 0.66in 0in 0in,clip,width=\textwidth]{figures/reconstruction.pdf}
%     \captionof{figure}{\textbf{Reconstruction quality comparison. } Our method can reconstruct out-of-domain samples very well (\ie the first two rows), while other methods fail to reconstruct due to blurry, wrong or completely face accessories (\ie hat, glasses). 
%     For less challenging inputs (\ie the last two rows), we show that the proposed method achieves better reconstruction quality with more accurate details (\ie hair style, cloth, eyebrows) than the other methods. 
%     Zoom-in recommended.
%     }
%     \label{fig:reconstruction}
%   \end{minipage}
%   \hfill
  \begin{minipage}[c]{0.48\textwidth}
        \begin{minipage}[c]{\textwidth}
        \centering
        \captionof{table}{\textbf{Quantitative evaluation on semantic face editing.} Results are reported with identity similarity (ID), L2 and LPIPS.}
        \begin{adjustbox}{max width=\textwidth}
            \begin{tabular}{cl|ccc}
            \toprule
            Encoder & Editing & ID ($\uparrow$) & L2 ($\downarrow$) & LPIPS ($\downarrow$) \\
            \midrule
            \parbox[t]{13mm}{\multirow{3}{*}{{pSp~\cite{richardson2021encoding}}}} & InterfaceGAN & 0.42 & 0.069 & 0.21\\ 
            \multirow{3}{*}{} & GANSpace & 0.50 & 0.111 & 0.24\\
            \multirow{3}{*}{} & SeFa & 0.53 & 0.078 & 0.22\\
            \midrule
            \parbox[t]{13mm}{\multirow{3}{*}{{e4e~\cite{tov2021designing}}}} & InterfaceGAN & 0.35 & 0.081 & 0.25\\ 
            \multirow{3}{*}{} & GANSpace & 0.42 & 0.114 & 0.27\\
            \multirow{3}{*}{} & SeFa & 0.44 & 0.107 & 0.25\\
            \midrule
            \parbox[t]{10mm}{\multirow{2}{*}{{Ours}}} & $edit$ & 0.52 & 0.049 & 0.17\\
            \multirow{2}{*}{} & $edit\_refine$ & 0.77 & 0.028 & 0.09\\
          \bottomrule
            \end{tabular}
            \end{adjustbox}
        \label{tbl:editing} 
        \end{minipage}
    \end{minipage}
    \hfill
    \begin{minipage}[c]{0.48\textwidth}
        \begin{minipage}[c]{\textwidth}
        \centering
        \captionof{table}{\textbf{Quantitative comparison on reconstructed images.} We report LPIPS, L2, PSNR and SSIM results on CelebA-HQ. We show consistent results with our qualitative evaluation as we achieve significantly better scores among all metrics.}
        \begin{adjustbox}{max width=\linewidth}
            \begin{tabular}{c|cccc}
            \toprule
            & LPIPS ($\downarrow$) & L2 ($\downarrow$) & PSNR ($\uparrow$) & SSIM ($\uparrow$) \\
            \midrule
            SG2~\cite{karras2020analyzing} & 0.31 & 0.038 & 19.03 & 0.558\\
            pSp~\cite{richardson2021encoding} & 0.16 & 0.036 & 16.42 & 0.513\\
            e4e~\cite{tov2021designing} & 0.19 & 0.048 & 15.13 & 0.490\\
            Ours & 0.07 & 0.014 & 20.41 & 0.834\\
            \bottomrule
        \end{tabular}
        \end{adjustbox}
        \label{tbl:reconstruction} 
        \end{minipage}
    \end{minipage}
\end{minipage}
\end{figure}

%% tab: editing
\begin{comment}
\begin{table}[t]
    \centering
    \begin{adjustbox}{max width=0.9\linewidth}
    \begin{tabular}{cl|ccc}
    \toprule
    Encoder & Editing & ID ($\uparrow$) & L2 ($\downarrow$) & LPIPS ($\downarrow$) \\
    \midrule
    \parbox[t]{13mm}{\multirow{3}{*}{{pSp~\cite{richardson2021encoding}}}} & InterfaceGAN & 0.42 & 0.069 & 0.21\\ 
    \multirow{3}{*}{} & GANSpace & 0.50 & 0.111 & 0.24\\
    \multirow{3}{*}{} & SeFa & 0.53 & 0.078 & 0.22\\
    \midrule
    \parbox[t]{13mm}{\multirow{3}{*}{{e4e~\cite{tov2021designing}}}} & InterfaceGAN & 0.35 & 0.081 & 0.25\\ 
    \multirow{3}{*}{} & GANSpace & 0.42 & 0.114 & 0.27\\
    \multirow{3}{*}{} & SeFa & 0.44 & 0.107 & 0.25\\
    \midrule
    \parbox[t]{10mm}{\multirow{2}{*}{{Ours}}} & $edit$ & 0.52 & 0.049 & 0.17\\
    \multirow{2}{*}{} & $edit\_refine$ & 0.77 & 0.028 & 0.09\\
  \bottomrule
    \end{tabular}
    \end{adjustbox}
    \caption{\textbf{Quantitative evaluation on semantic face editing.} Results are reported with identity similarity (ID), L2 and LPIPS.}
    \small
    \label{tbl:editing} 
\end{table}
\end{comment}

% \subsubsection{Quantitative evaluation}
\noindent\textbf{Quantitative evaluation.}
We then quantitatively compare our approach with various editing methods. Again, we first estimate the latent codes of images using both pSp and e4e encoders, then apply editing methods on these latent codes to synthesize editing images.
As shown in Table~\ref{tbl:editing}, we achieve better performance with significantly higher identity similarity between the source and edited images. 
Besides, our approach has lower L1 and LPIPS scores, which shows that our method encourages local and disentangled manipulation of desired attributes, while other methods may include some unexpected changes of other semantics. Average results of four attributes (i.e. glass, age, smiling, and gender) are reported.

\subsection{Real Face Reconstruction}
We then compare the proposed method to the current inversion methods (\ie SG2~\cite{karras2020analyzing}, pSp~\cite{richardson2021encoding}, and e4e~\cite{tov2021designing}) in terms of reconstruction quality.
SG2 is an optimization-based GAN inversion method that inverts real images to the latent space of StyleGAN, while pSp and e4e are encoder-based inversion approaches.
Note that real face images tend to be out of the distributions of StyleGAN latent space in terms of appearance (\eg occlusions, extreme lighting conditions, detailed hairstyles). To mitigate the inversion distortion and achieve higher reconstruction quality, we invert real images to the $\mathcal{W+}$ space for all methods.
% the extended $\mathcal{W+}$ space 

% \subsubsection{Qualitative evaluation}
\noindent\textbf{Qualitative evaluation.}
Figure~\ref{fig:reconstruction} shows the qualitative comparison of different methods on reconstructed images.
As can be seen, SG2~\cite{karras2020analyzing} achieves fairly well reconstruction results. However, our method shows significantly better results on out-of-domain samples (\ie the first two rows of Figure~\ref{fig:reconstruction}). The proposed method is the only one that can accurately reconstruct challenging face accessories (\ie hat, glasses that are not placed on the eyes).
Besides, our method is more efficient compared to SG2 as it is an optimization-based method.
Note that we do not perform style editing on the latent code estimated by SG2, as we observe that it can not generate significant edits.
% Note that we do not include SG2 for latent code estimating, as we observe that it can not generate significant edits.
Compared to the other two encoder-based methods (\ie pSp~\cite{richardson2021encoding} and e4e~\cite{tov2021designing}), we achieves superior results on all examples with sharper edges and finer details (\ie hairstyle, facial expression, illumination, and wrinkles).
More comparisons can be found in the Supplementary Material.

%% tab: reconstruction
\begin{comment}
\begin{table}[t]
    \centering
    \begin{adjustbox}{max width=0.9\linewidth}
    \begin{tabular}{c|cccc}
    \toprule
    & LPIPS ($\downarrow$) & L2 ($\downarrow$) & PSNR ($\uparrow$) & SSIM ($\uparrow$) \\
    \midrule
    SG2~\cite{karras2020analyzing} & 0.31 & 0.038 & 19.03 & 0.558\\
    pSp~\cite{richardson2021encoding} & 0.16 & 0.036 & 16.42 & 0.513\\
    e4e~\cite{tov2021designing} & 0.19 & 0.048 & 15.13 & 0.490\\
    Ours & 0.07 & 0.014 & 20.41 & 0.834\\
  \bottomrule
    \end{tabular}
    \end{adjustbox}
    \caption{\textbf{Quantitative comparison on reconstructed images.} We report LPIPS, L2, PSNR and SSIM results on CelebA-HQ. We show consistent results with our qualitative evaluation as we achieve significantly better scores among all metrics.}
    \small
    \label{tbl:reconstruction} 
\end{table}
\end{comment}

% \subsubsection{Quantitative evaluation}
\noindent\textbf{Quantitative evaluation.}
We then quantitatively evaluate the performance of our method in comparison to other methods on reconstructed images. 
As shown in Table~\ref{tbl:reconstruction}, we report the perceptual similarity (\ie LPIPS~\cite{zhang2018unreasonable}), the pixel-wise distance (\ie L2), the peak signal to noise ratio (\ie PSNR), and the structural similarity (\ie SSIM~\cite{wang2003multiscale}) between the source and reconstructed images. Quantitative results show that our method achieves the best among all metrics by a significant margin, which is also aligned with qualitative evaluations.

%% fig: ablation_l1_reg2
\begin{comment}
\begin{figure}[t]
	\centering   
    \includegraphics[trim=0in 0.2in 0in 0in,clip,width=0.5\linewidth]{figures/ablation_l1_reg.pdf}
    \caption{\textbf{Ablation study on sparsity regularization.} By adopting sparsity regularization on editing directions, we can achieve disentangled and local semantic manipulations. In the first column, we can see that the glass and age attributes are more disentangled. Moreover, sparsity regularization also helps remove unexpected attributes changing (\eg face color and hairstyles).}\label{fig:ablation_l1_reg}
\end{figure}

\end{comment}

%% fig: ablation_l1_reg2
%% tab: ablation
%% fig: ablation_refine
%% fig: ablation_align_loss
\begin{figure}[!t]
\begin{minipage}{\linewidth}
    \begin{minipage}[c]{0.48\textwidth}
        \begin{minipage}[c]{\textwidth}
        \centering
        \captionof{table}{\textbf{Quantitative evaluation for ablation study.} Results are reported with identity similarity (ID), L2 and LPIPS.
        }
        \begin{adjustbox}{max width=\textwidth}
            \begin{tabular}{cccc|ccc}
            \toprule
            %  & ID ($\uparrow$) & L2 ($\downarrow$) & LPIPS ($\downarrow$) \\
            % \midrule
            % w/o $\mathcal{L}_{spa\_reg}$ & 0.42 & 0.069 & 0.21\\ 
            % w/o $\mathcal{L}_{align}$ & 0.50 & 0.111 & 0.24\\
            % w/o refinement branch & 0.50 & 0.111 & 0.24\\
            % Ours & 0.53 & 0.078 & 0.22\\
            S-branch & $\mathcal{L}_{spa}$ & R-branch & $\mathcal{L}_{align}$ & ID ($\uparrow$) & L2 ($\downarrow$) & LPIPS ($\downarrow$) \\
            \midrule
            \checkmark&&& & 0.34 & 0.106 & 0.24\\ 
            \checkmark&\checkmark&& & 0.52 & 0.049 & 0.17\\
            \checkmark&\checkmark&\checkmark& & 0.69 & 0.059 & 0.15\\
            \checkmark&\checkmark&\checkmark&\checkmark & 0.77 & 0.028 & 0.09\\
          \bottomrule
        \end{tabular}
        \end{adjustbox}
        \label{tbl:ablation}
      \end{minipage}
      
        \vspace{3mm}
        \begin{minipage}[c]{\textwidth}
        \centering
        \includegraphics[trim=0in 0.2in 0in 0in,clip,width=\linewidth]{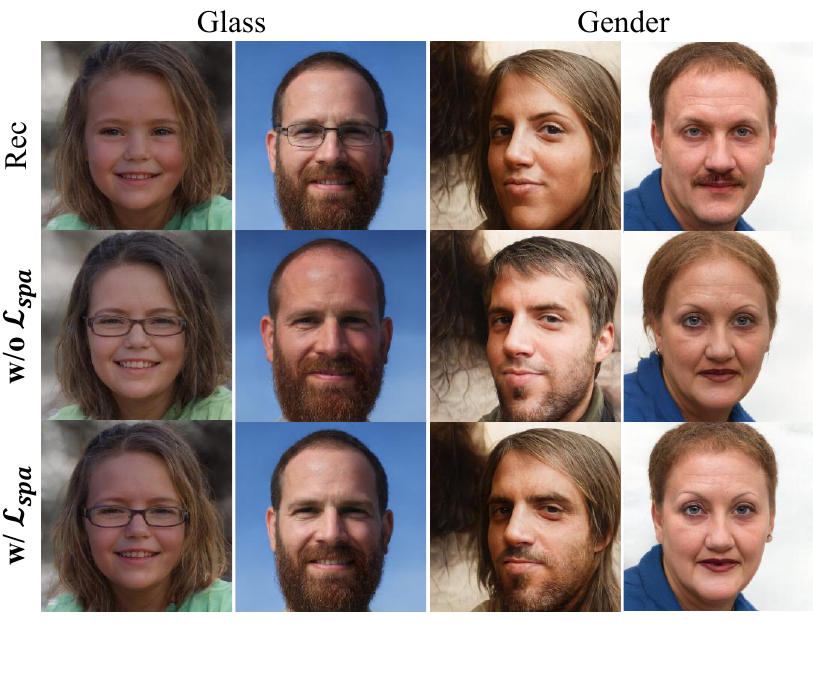}
        \captionof{figure}{\textbf{Ablation study on sparsity regularization.} By adopting sparsity regularization on editing directions, we can achieve disentangled and local semantic manipulations. In the first column, we can see that the glass and age attributes are more disentangled. Moreover, sparsity regularization also helps remove unexpected attributes changing (\eg face color and hairstyles).
        }
        \label{fig:ablation_l1_reg}
      \end{minipage}
    \end{minipage}
    \hfill
    \begin{minipage}[c]{0.48\textwidth}
        \begin{minipage}[c]{\textwidth}
        \centering
        \includegraphics[trim=0in 0in 0in 0in,clip,width=\textwidth]{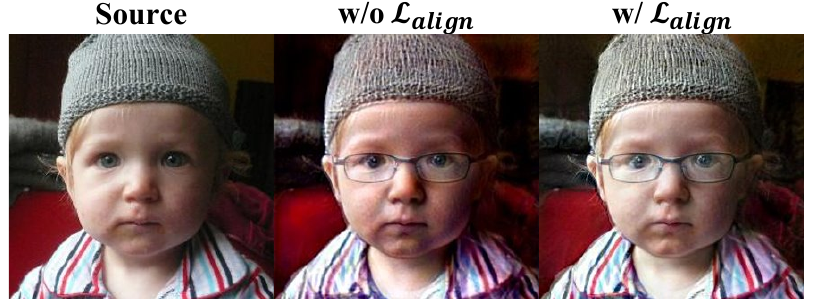}
        \captionof{figure}{\textbf{Effect of the alignment loss.} $\mathcal{L}_{align}$ provides strong supervision for the training of our model. Removing $\mathcal{L}_{align}$ will lead to color inaccuracy.}
        \label{fig:ablation_align_loss} 
        \end{minipage}
                
        \vspace{3mm}
        \begin{minipage}[c]{\textwidth}
        \centering
        \includegraphics[trim=0in 0in 0in 0in,clip,width=\textwidth]{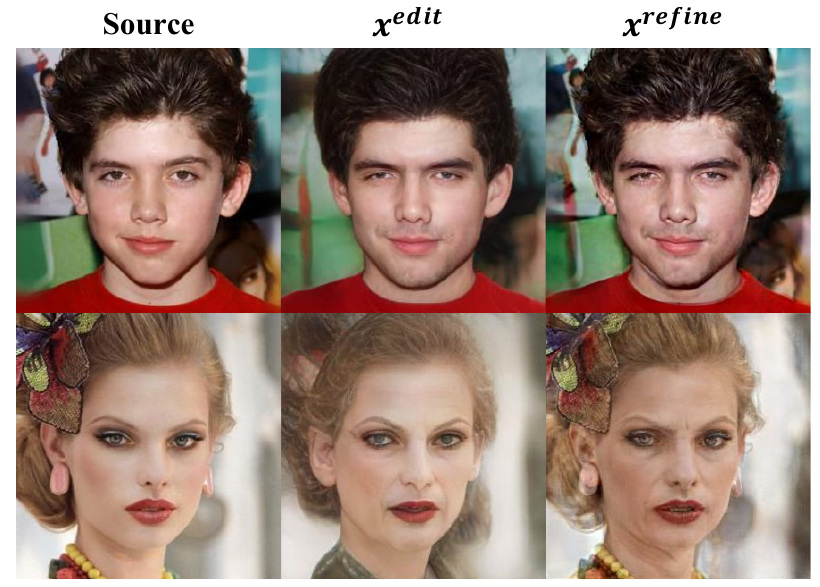}
        \captionof{figure}{\textbf{Effect of expanding the latent space with extra texture information.} We show results of $\x^{edit}$ and $\x^{ref\_edit}$ on the gender (\ie top row) and age (\ie bottom row) editing. Before expanding the latent space, the proposed method achieves reasonable and disentangled editing (\eg $\x^{edit}$). Then the edits are further refined with more details  provided by the 2d content features.}
        \label{fig:ablation_refine} 
        \end{minipage}
    \end{minipage}
\end{minipage}
\end{figure}

\subsection{Ablation Study and Discussion}
We conduct ablation studies to justify our model designs. 
Quantitative results for ablation study is shown in Table~\ref{tbl:ablation}.

\noindent\textbf{Sparsity regularization for disentangled editing.}
We show the effect of sparsity regularization in Figure~\ref{fig:glass_dir_heatmap} and~\ref{fig:ablation_l1_reg}. The visual results demonstrate that only a limited number of elements in latent space need to be changed to perform semantic editing with high quality and untangling properties.
Moreover, sparsity regularization also helps remove unexpected attributes changing. As shown in Figure~\ref{fig:ablation_l1_reg}, editing along glass or gender will not change the face color or hairstyles in the source image.

\noindent\textbf{Texture consultation.}
The style editing branch of our model aims to search for the disentangled editing direction. As can be seen in Figure~\ref{fig:ablation_refine}, $\x^{edit}$ achieves reasonable edits, but the distortion is large.
The content refinement branch uses additional content and texture details obtained from the source image to reduce edit distortion and generate fine edits.
An alignment loss is introduced to successfully train the content refinement branch. We show its effect in Figure~\ref{fig:ablation_align_loss}.

\noindent\textbf{Limitation.}
The proposed model includes two branches to tackle the entanglement and distortion of semantic manipulation, respectively. The sparsity regularization in the first branch is an essential part in our model. 
Not only is it helpful to eliminate unexpected attributes changes, the calculation of the relevance map and the alignment loss are also highly dependent on the spatial disentanglement performance of the edits.
However, we discovered a side effect of the sparsity regularization, that is, it limits the editing of large spatial changes, such as pose editing. One explanation is that pose editing requires to change a large region of the image, and it is too complicated to be achieved by changing a sparse matrix on the latent code.

% We found that pose is well disentangled in $\mathcal{W}$ space, which demonstrates that the spatial change is related to the sparsity of editing directions. 

% The goal of the first branch is to search for disentangled semantic editing directions, 
% The second branch of our model aims to reduce the editing distortion and generate refined edits with more texture details of the source image.
% It has been shown that using less editable embedding spaces, such as W+, results in better reconstruction, but also in less meaningful editing compared to the native W space.